\documentclass[letterpaper]{article} 
\usepackage{aaai25}  
\usepackage{times}  
\usepackage{helvet}  
\usepackage{courier}  
\usepackage[hyphens]{url}  
\usepackage{graphicx} 
\usepackage{natbib}  
\usepackage{caption} 
\usepackage{algorithm}
\usepackage{algorithmic}
\usepackage{booktabs}
\usepackage{multicol}
\usepackage{multirow}
\usepackage{amssymb}
\usepackage{amsmath}
\usepackage{subfigure}

\newcommand{\mysubscript}[1]{\raisebox{1.8pt}{\textsubscript{#1}}}
\newcommand{\model}{SEPC}
\title{Structural Entropy Guided Probabilistic Coding}
\author{
    Xiang Huang\textsuperscript{\rm 1},
    Hao Peng\textsuperscript{\rm 1,\rm 2}\thanks{Corresponding author.},
    Li Sun\textsuperscript{\rm 3},
    Hui Lin\textsuperscript{\rm 4},
    Chunyang Liu\textsuperscript{\rm 5},
    Jiang Cao\textsuperscript{\rm 6},
    Philip S. Yu\textsuperscript{\rm 7}
}
\affiliations{
    \textsuperscript{\rm 1}Beihang University,\\
    \textsuperscript{\rm 2}Guangdong Laboratory of Artificial Intelligence and Digital Economy\\
    \textsuperscript{\rm 3}North China Electric Power University\\
    \textsuperscript{\rm 4}China Academic of Electronics and Information Technology\\
    \textsuperscript{\rm 5}Didi Chuxing\\
    \textsuperscript{\rm 6}Academy of Military Sciences\\
    \textsuperscript{\rm 7}University of Illinois Chicago\\
    \{huang.xiang, penghao\}@buaa.edu.cn,
    ccesunli@ncepu.edu.cn,
    linhui@cetc.com.cn\\
    liuchunyang@didiglobal.com,
    amscaojiang@126.com,
    psyu@uic.edu
}

\begin{document}

\maketitle

\begin{abstract}
Probabilistic embeddings have several advantages over deterministic embeddings as they map each data point to a distribution, which better describes the uncertainty and complexity of data. 
Many works focus on adjusting the distribution constraint under the Information Bottleneck (IB) principle to enhance representation learning.
However, these proposed regularization terms only consider the constraint of each latent variable, omitting the structural information between latent variables.
In this paper, we propose a novel structural entropy-guided probabilistic coding model, named \model{}.
Specifically, we incorporate the relationship between latent variables into the optimization by proposing a structural entropy regularization loss.
Besides, as traditional structural information theory is not well-suited for regression tasks, we propose a probabilistic encoding tree, transferring regression tasks to classification tasks while diminishing the influence of the transformation. 
Experimental results across 12 natural language understanding tasks, including both classification and regression tasks, demonstrate the superior performance of \model{} compared to other state-of-the-art models in terms of effectiveness, generalization capability, and robustness to label noise.
The codes and datasets are available at \url{https://github.com/SELGroup/SEPC}.
\end{abstract}

\section{Introduction}\label{sec:intro}
Probabilistic embedding~\cite{vilnis2015word} is a flexible representation learning method aiming to learn the underlying probability distribution of data.
It has been broadly applied to various domains such as graph structural learning~\cite{sun2022graph}, computer vision~\cite{kim2021dropbottleneck, oh2019modeling, shi2019probabilistic, fischer2020conditional}, and natural language processing~\cite{mahabadi2021variational,hu2024structured, hu2022varmae}.
In contrast to deterministic embedding~\cite{Dong2024, xu2024sctnet}, which maps the input into a fixed latent variable representation, probabilistic embedding represents each data point as a probability distribution.
Hence, probabilistic embedding inherently accounts for the uncertainty and complexity of data by controlling the spread of the probability density over the learning latent space~\cite{oh2019modeling}, showcasing better discriminative ability and robustness.

The mainstream probabilistic embedding methods are grounded in the Information Bottleneck (IB) principle~\cite{tishby2000information, tishby2015deep}.
IB aims to find compressed representations that maintain as much information as possible for the prediction task while removing as much irrelevant information as possible.
Specifically, it seeks the latent representation $Z$ that is maximally informative about the target $Y$ (i.e., maximize mutual information $I(Y; Z)$) while being minimally informative about the input data $X$ (i.e., minimize mutual information $I(X; Z)$)~\cite{sun2022graph}.
The former target is typically achieved with common task losses like cross entropy (CE) loss or mean squared error (MSE) loss, whereas various regularization losses are proposed for the latter goal.
VIB~\cite{dvib} assumes the prior distribution of $Z$ is the standard normal distribution and utilizes Kullback–Leibler (KL) divergence to regularize the learning distribution $p(z|x)$.
Sparse IB~\cite{chalk2016relevant} changes the prior distribution of VIB to the Student-t distribution to achieve relevant and sparse coding.
MEIB~\cite{an2023maximum} lifts the prior distribution constraint of VIB and instead uses maximum conditional entropy $H(Z|X)$ as the only regularization.
SPC~\cite{hu2024structured} omits the decoder of VIB and proposes an additional structured regularization that encourages class-level uniformity within the latent space under the multivariate Gaussian distribution.
However, all of them focus solely on the individual latent variable $Z$ or the constraint of $Z$ with the label $Y$, neglecting the structural information between latent variables.

In recent years, structural entropy theory~\cite{li2016structural} has demonstrated its advantage in capturing hierarchical structural information and has been widely used in various fields like node classification~\cite{duan2024structural}, graph structural learning~\cite{zou2023segsl}, and contrastive learning~\cite{wu2023sega}.
It considers the structural information of the original inputs by modeling the input data as a graph and then converting the graph into an encoding tree.
The data points are the leaf nodes of the encoding tree, and each upper node represents a partition, resulting in a hierarchical clustering of the input data.
Low-depth tree nodes depict more coarse-grained clusters of the input data.
Each node in the encoding tree has its own structural entropy.
The structural entropy of the encoding tree is calculated by summing the structural entropy of all non-root nodes, representing the overall structural information of the input.
Previous work~\cite{wang2023user, zeng2023effective} mostly focuses on minimizing the structural entropy of the encoding trees to obtain the optimized encoding tree or embeddings of input data, aiming at learning as much task-related information as possible.
However, the potential for using structural entropy for regularization remains underexplored.
Additionally, as the structural entropy is designed for classification tasks, how to effectively leverage it in the regression task is still a problem.

In this paper, we propose a structural entropy-guided probabilistic coding model, named \model{}.
We present a structural entropy-based regularization loss that incorporates structural information between latent variables.
Specifically, we first construct the adjacency matrix based on the similarity between embeddings of latent variables and propose to maximize the structural entropy of the induced graph, which helps improve the generalization of the model by separating the probabilistic distribution of each latent variable.
Additionally, we design a probabilistic encoding tree to adapt our structural entropy loss in regression tasks.
We first discretize and soften regression labels into soft classification labels (i.e., each data point belongs to multiple classes with varying probabilities), diminishing the influence of unsuitable classification caused by using only discretization~\cite{pintea2023step}.
To adapt structural entropy to such soft labels, we relax the constraint that one child belongs to one parent in the encoding tree, allowing each child to connect to all upper-level nodes with varying probabilities.
Extensive experiments are conducted on 12 natural language understanding tasks, including 10 classification tasks and 2 regression tasks.
Comparative results and analysis demonstrate that the proposed \model{} enjoys superior effectiveness, generalization, and robustness compared to the state-of-the-art (SOTA) baselines.
The main contributions are summarized as follows:

\begin{itemize}
\item We present a structural entropy based regularization loss, incorporating the structural information between data points into model regularization. To our knowledge, this is the first time that maximizing structural entropy has been utilized as a regularization loss.

\item We propose a probabilistic encoding tree for soft classification labels and present an effective method to utilize structural entropy for regression tasks for the first time.

\item Extensive experiments on 12 datasets demonstrate that \model{} achieves SOTA performance in classification and regression tasks regarding effectiveness, generalization, and robustness.
\end{itemize}
\section{Preliminaries}\label{sec:priliminaries}

In this section, we present the basic concepts of probabilistic coding, the encoding tree, and structural entropy.

\subsection{Probabilistic Coding}
\begin{figure}[t]
    \centering
    \subfigure[Encoder-Decoder architecture.]{
        \label{fig:framework1}
        \begin{minipage}[t]{0.9\linewidth}
            \centering        \includegraphics[width=1\linewidth]{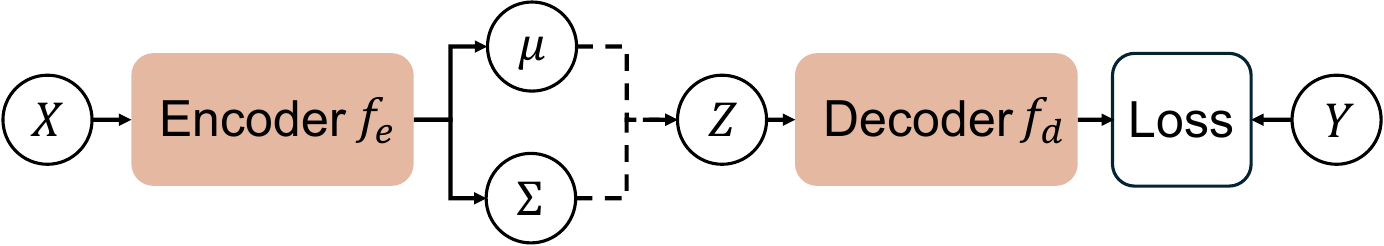}
        \end{minipage}%
    }
    \subfigure[Encoder-only architecture.]{
        \label{fig:framework2}
        \begin{minipage}[t]{0.9\linewidth}
            \centering        \includegraphics[width=1\linewidth]{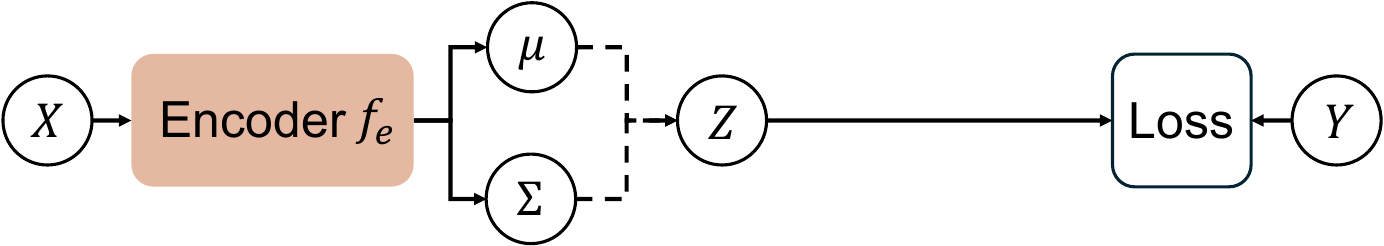}
        \end{minipage}%
    }
    \caption{Two common architectures of probabilistic coding.}
    \label{fig:framework}
\end{figure}

The classical probabilistic coding model employs an encoder-decoder architecture, as shown in Figure~\ref{fig:framework1}.
The encoder $f_e$ maps input $x \in X$ to a Gaussian distribution $\mathcal{N}(z;\mu, \Sigma)$.
All distributions of $z$ consist of the embedding space of the latent variable $Z$.
The re-parameterization trick~\cite{kingma2013autoencoding} is then used to sample $z$ from the distribution while keeping the gradient unbiased.
Finally, $z$ is mapped by the decoder to $f_d(z)$ to predict the label $y \in Y$.
The work~\cite{hu2024structured} also proposes an encoder-only architecture for probabilistic coding (as shown in Figure~\ref{fig:framework2}), omitting the decoder and directly predicting $y$ using the sample from the learned distribution.

Under the Markov chain constraint $Y\rightarrow X \rightarrow Z$, 
the probabilistic coding follows the Information Bottleneck principle and aims to learn the minimal sufficient information for representation $Z$:
\small
\begin{equation}
    Z = \mathop{argmin}\limits_{Z} -I(Z;Y) + \beta I(Z;X),
    \label{eq:IB}
\end{equation}    
\normalsize
where $I(Z;Y)$ is the mutual information between $Z$ and $Y$, $I(Z;X)$ is the mutual information between $Z$ and $X$, and $\beta$ is the Lagrangian multiplier trading off sufﬁciency and minimality.
Assuming $z\in Z$ follows the Gaussian distribution, the objective of probabilistic coding is as follows:
\small
\begin{equation}
    \mathcal{L}_{PC} = \mathbb{E}_{z\sim p(z|x)}\left[-\log q(y|z) \right]+ \beta \mathrm{KL}\left[p(z|x), r(Z)\right].
    \label{eq:PC loss}
\end{equation}
\normalsize
Here, $\mathrm{KL}$ refers to the KL divergence operator, $p(z|x) = \mathcal{N}(z;\mu, \Sigma)$ is learned by the encoder $f_e$, $r(Z)$ is the expected prior distribution, and $r(Z) = \mathcal{N}(z; \mathbf{0}, \mathbf{I})$ in general.
$q(y|z)$ is the variational approximation to $p(y|z)$ and is calculated by the decoder $f_d$ or by a non-parametric operator like the softmax function in the encoder-only architecture~\cite{hu2024structured}.

\subsection{Encoding Tree}
Given a graph $G = \{X, E, W\}$, $X$ is the set of input data points, $E$ is the edge set, and $W\in \mathbb{R}^+$ is the edge weight set.
For each point $x\in X$, its degree $d_x$ is defined as the sum of the weights of edges associated with it.
The encoding tree $\mathcal{T}$ of $G$ is a multi-child tree with the following properties:
(1) Each tree node $\alpha$ corresponds to a subset of data points $T_{\alpha}\subseteq X$. 
Especially, for the root node $\lambda$ of $\mathcal{T}$, we define the points set it associated with as $T_{\lambda} = X$.
For the leaf node $\alpha$ at the last depth, $T_{\alpha}$ is a singleton containing a single data point $x \in X$.
If the leaf node $\alpha$ is not at the last depth, $T_{\alpha}$ is $\emptyset$.
(2) For each non-leaf tree node $\alpha$, its $i$-th immediate child is $\alpha^{<i>}$, and its parent node is denoted as $\alpha^-$.
(3) For each non-leaf tree node $\alpha$, $T_{\alpha} = \bigcup_{i=1}^{N_\alpha} T_{\alpha^{<i>}}$, $N_{\alpha}$ is the number of children of $\alpha$.
With these properties, each depth of a node in the encoding tree depicts a partition of the data point set $X$, and lower depth means a more coarse-grained partition.

\subsection{Structural Entropy}
The structural entropy is defined under the graph $G$ and the encoding tree $\mathcal{T}$ as follows:
\small
\begin{align}
    H^{\mathcal{T}}(G) &= \sum_{\alpha\in \mathcal{T}, \alpha\neq \lambda} H^{\mathcal{T}}(G; \alpha),\\
    H^{\mathcal{T}}(G; \alpha) &= -\frac{g_{\alpha}}{\mathrm{vol}(G)}\log_2\frac{\mathcal{V}_{\alpha}}{\mathcal{V}_{\alpha^-}}.  
\end{align}
\normalsize
Here, $g_{\alpha}$ is the sum of the weights of the edges that connect points inside $T_{\alpha}$ with points outside $T_{\alpha}$ (i.e., the weights of the cut edges between $T_{\alpha}$ and its complement set $T_{\alpha}^\complement$).
The volume of $G$, denoted as $\mathrm{vol}(G)$, is the sum of the degrees of all data points $X$, i.e., $\mathrm{vol}(G) = \sum_{x \in X} d_x$.
$\mathcal{V}_\alpha = \sum_{x\in T_{\alpha}} d_x$ is the volume of $T_{\alpha}$, and $\alpha^-$ is the parent node of $\alpha$. 
\section{Proposed Method}\label{sec:method}
\begin{figure}[t]
    \centering
    \includegraphics[width=0.9\linewidth]{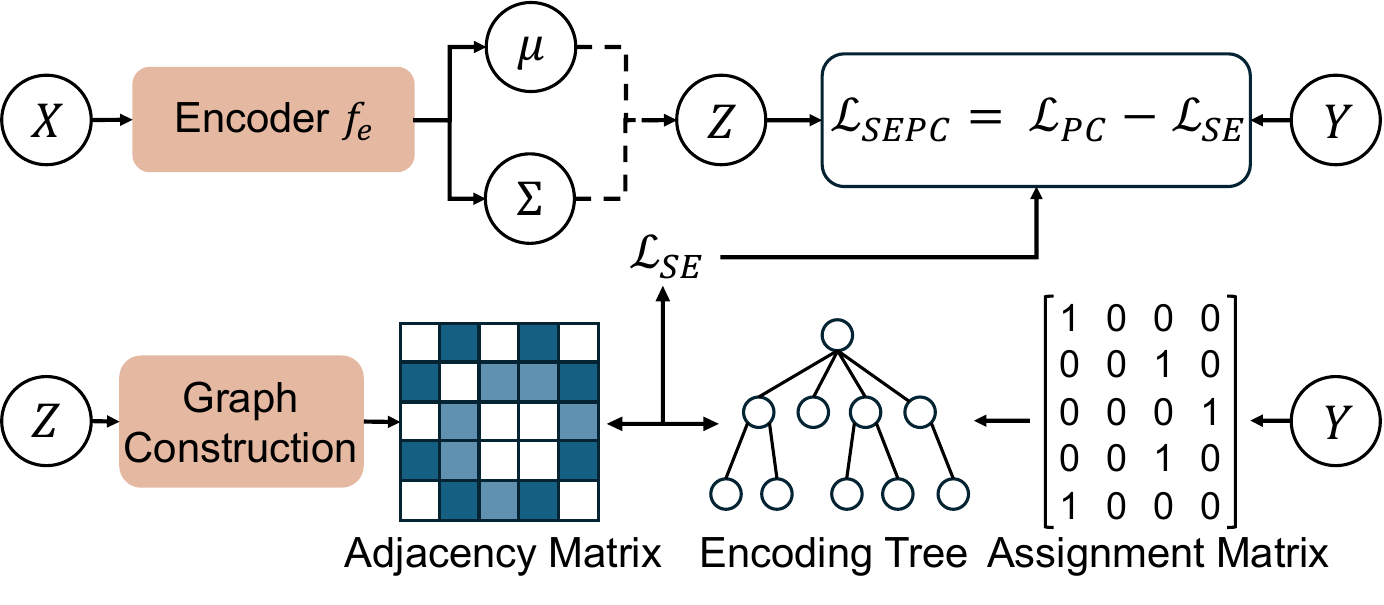}
    \caption{The overall model of \model{}.}
    \label{fig:sepc}
\end{figure}
In this section, we elaborate on the proposed structural entropy based regularization loss of \model{}, introduce the probabilistic encoding tree for soft classification labels, and describe how to utilize it in regression tasks.
We adopt the encoder-only architecture~\cite{hu2024structured} for probabilistic coding, and the overall model of \model{} is shown in Figure~\ref{fig:sepc}.

\subsection{Structural Entropy based Regularization Loss}
Previous works only consider the individual latent variable in the regularization loss, ignoring the structural information between latent variables.
To capture the structural information, we incorporate structural entropy into the regularization loss, as it inherently considers the self-organization of data.
As illustrated in Figure~\ref{fig:sepc}, the input data $X$ is first encoded into the probabilistic embedding $H_Z$.
The graph $G$ is constructed from $Z$ as follows:
\small
\begin{equation}
    \label{eq:graph construction}
    A = \sigma(H_Z \times H_Z^T),
\end{equation}
\normalsize
where $H_Z$ is the embedding of $Z$, and $\sigma$ is the sigmoid activation function to ensure positive values for the adjacency matrix $A$.

The construction of the encoding tree is also straightforward.
We treat the labels as the optimal partition for the data and construct a three-tier encoding tree.
The nodes in the intermediate layer represent the classes of the classification task, and each leaf node (i.e., the input data $X$) is assigned to an intermediate node according to its label.
We define an assignment matrix $C\in \{0, 1\}^{n\times r}$, where $n$ is the number of leaf nodes and $r$ is the number of intermediate nodes.
$C_{ij} = 1$ means the $i$-th leaf node belongs to the $j$-th class.
To enhance the capability of the latent representations, we propose maximizing the structural entropy of the intermediate layer nodes, constraining the probabilistic distribution of the latent variables to ensure separation.
The structural entropy of the intermediate layer nodes for the three-tier encoding tree is as follows:
\small
\begin{equation}
    \label{eq:intermediate nodes se}
    H^{\mathcal{T}}_{C}(G) = \sum_{j=1}^{r}\frac{-g_{\alpha_j}}{\mathrm{vol}(G)}\log_2 \frac{\mathcal{V}_{\alpha_j}}{\mathrm{vol}(G)},
\end{equation}
\normalsize
where $r$ is the number of classes, $g_{\alpha_j}$ is the sum of the weights of the cut edges between $T_{\alpha_j}$ and its complement set $T_{\alpha_j}^\complement$, $\mathcal{V}_{\alpha_j}$ is the volume of $T_{\alpha_j}$, and $\{\alpha_1, \dots, \alpha_r\}$ is the intermediate layer nodes in the encoding tree.
Utilizing the adjacency matrix $A$ and the assignment matrix $C$, the regularization loss format of $H^{\mathcal{T}}_{C}(G)$ is as follows:
\small
\begin{equation}
    \label{eq:se loss}
    \mathcal{L}_{SE} = -\sum_{j=1}^{r}\frac{\left((\mathbf{1}-C)^TAC\right)_{jj}}{\mathrm{sum}(A)}\times \log_2\frac{\left(\mathbf{1}^TAC\right)_{jj}}{\mathrm{sum(A)}}.
\end{equation}
\normalsize
Here, $\mathbf{1}$ is the full-one matrix with shape $n\times r$, the operator $\mathrm{sum}(\cdot)$ sums up the matrix to a scalar, and $(\cdot)_{jj}$ selects the value in the $j$-th row and the $j$-th column of the matrix.
The overall loss of \model{} is as follows:
\small
\begin{equation}
    \label{eq:sepc loss}
    \mathcal{L}_{SEPC} = \mathcal{L}_{PC} - \gamma\mathcal{L}_{SE},
\end{equation}
\normalsize
where $\gamma$ is a hyperparameter controlling the weight of our structural entropy based regularization loss $\mathcal{L}_{SE}$.

\begin{figure}[t]
    \centering
    \includegraphics[width=0.9\linewidth]{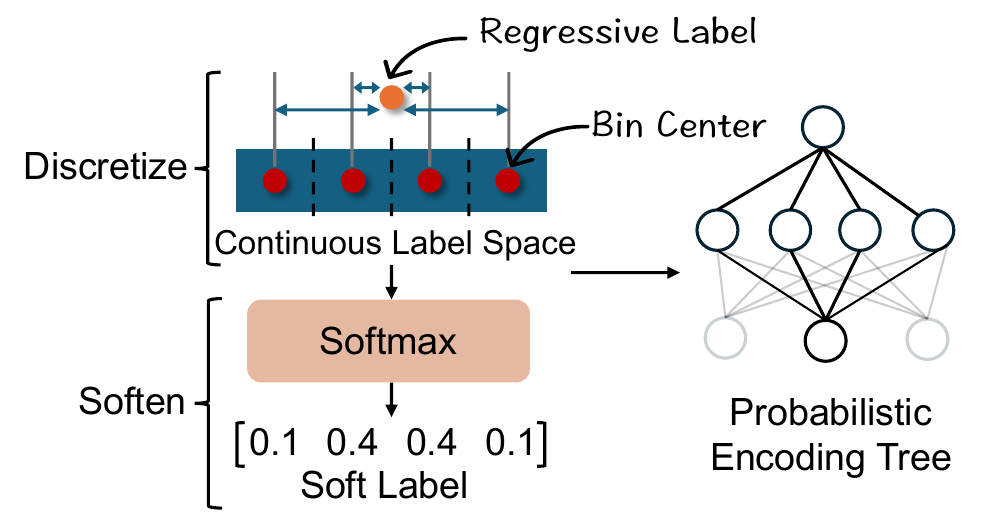}
    \caption{Probabilistic encoding tree for regression tasks.}
    \label{fig:probabilistic encoding tree}
\end{figure}

\subsection{Probabilistic Encoding Tree for Regression Tasks}

Discretization is a widely used method to transform a regression task into a classification task by binning continuous labels into discrete classes~\cite{muthukumar2021classification, stewart2023regression}.
However, as the binning borders need to be predefined, inappropriate borders can lead to unbalanced or indistinguishable classification labels, hampering model performance~\cite{pintea2023step}. 
Softening labels mitigates this issue ~\cite{Ma2023EnhancedSL}, as it allows each data point to belong to all classes with different probabilities to express tendencies.
We propose a probabilistic encoding tree to utilize structural entropy theory in such soft classification labels. 
It loosens the constraint that one child node is only assigned to one parent node, allowing the child node to connect with all up-depth nodes with different probabilities.

As shown in Figure~\ref{fig:probabilistic encoding tree}, during the discretized period, we first bin the entire regression label value space into $r$ classes.
Then, we calculate the distance between the regressive label $Y$ and the centers $P = \{P_1, \dots, P_r\}$ of the $r$ bins:
\small
\begin{equation}
    D = |Y^T - P|,
\end{equation}
\normalsize
where $D\in \mathbb{R}^{n\times r}$, $n$ is the number of data points, and the $i$-th row of $D$ denotes the distance between the $i$-th data point and the $r$ bin centers.
The soft label is then calculated during the softening period as follows:
\small
\begin{equation}
    Y' = \mathrm{softmax}(-D),
\end{equation}
\normalsize
where $-D$ ensures that a closer distance to the bin center corresponds to a higher probability of belonging to this class.
The structural entropy of the intermediate layer nodes for the three-tier probabilistic encoding tree $H_{C}^{\mathcal{T}}(G)$ is then defined as follows:
\small
\begin{align}
\label{eq:probabilistic se}
    \mathcal{V}'_{\alpha_j} &= \sum_{x_i\in X} Y_{ij}' d_{x_i},\\
    H^{\mathcal{T}}(G;\alpha_j) &= -\frac{g_{\alpha_j}'}{\mathrm{vol}(G)} \log_2 \frac{\mathcal{V}_{\alpha_j}'}{\mathrm{vol}(G)},\\
    H_{C}^{\mathcal{T}}(G) &= \sum_{j=1}^r H^{\mathcal{T}}(G; \alpha_j).
\end{align}
\normalsize
Here, $Y_{ij}'$ denotes the soft label of the $i$-th data point $x_i$ regarding to the $j$-th class, and $d_{x_i}$ is the degree of $x_i$.
$\alpha_j$ represents the $j$-th intermediate layer nodes in the probabilistic encoding tree.
For $g_{\alpha_j}'$, the weight of cut edges should be multiplied by the probability of one vertex belonging to $T_{\alpha_j}$ and the other belonging to $T_{\alpha_j}^\complement$.
Letting the assignment matrix $C = Y'$, the structural entropy loss for the probabilistic encoding tree is as follows:
\small
\begin{equation}
    \label{eq:probabilistic se loss}
    \mathcal{L}_{SE} = -\sum_{j=1}^{r}\frac{\left((\mathbf{1}-C)^TAC\right)_{jj}}{\mathrm{sum}(A)}\times \log_2\frac{\left(\mathbf{1}^TAC\right)_{jj}}{\mathrm{sum(A)}}.
\end{equation}
\normalsize
It is equivalent to Equation~\ref{eq:se loss} in the formula, except that the elements of the assignment matrix $C$ are probabilities between $0$ and $1$.
Thus far, we have presented an effective method to utilize structural entropy for regression tasks.

\section{Experiments}\label{sec:experiments}
In this section, we conduct extensive experiments to evaluate the effectiveness, generalization capability, and robustness of \model{}. For fairness, all results are reported as the average and standard deviation of metrics tested with five random seeds, as in other works.

\subsection{Experiment Setups}
\subsubsection{Datasets}
Following \citet{hu2024structured}, we evaluate \model{} on 10 classification task datasets and 2 regression task datasets.
For classification tasks, 7 datasets about tweet semantic analysis are used: Emoji~\cite{barbieri2018semeval}, Emotion~\cite{mohammad2018semeval}, Hate~\cite{basile2019semeval}, Irony~\cite{van2018semeval}, Offensive~\cite{zampieri2019semeval}, Sentiment~\cite{rosenthal2019semeval}, and Stance~\cite{mohammad2016semeval}.
Additionally, we also experiment on three emotion-related datasets from different domains: ISEAR~\cite{scherer1994evidence}, MELD~\cite{poria-etal-2019-meld}, and GoEmotions~\cite{demszky-etal-2020-goemotions}.
For regression tasks, we utilize STS-B~\cite{cer-etal-2017-semeval} and Claire~\cite{roth-etal-2022-semeval} for evaluation.

\subsubsection{Evaluation Metric}
We use the same metric as in previous works.
The macro-averaged F1 score across all classes is reported for most classification datasets.
Following \citet{hu2024structured}, we report the macro-averaged F1 score of favor and against classes for the Stance dataset, the F1 score of the ironic class for the Irony dataset, and the macro-averaged recall for the Sentiment dataset.
For regression tasks, we report both Pearson and Spearman correlation coefficients.

\subsubsection{Baselines}
We compare \model{} with two categories of classic baselines: universal models and fine-tuned representation models.
The baseline results are collected from the work of ~\citeauthor{hu2024structured} or evaluated using the source code provided by the authors.
In the universal models, we compare with SVM~\cite{cortes1995support}, FastText~\cite{joulin-etal-2017-bag}, BiLSTM~\cite{hochreiter1997long}, and GPT-3.5\footnote{https://openai.com/index/chatgpt/}.
For the fine-tuned models, we use bert-base-uncased~\cite{devlin2019bert} and roberta-base~\cite{liu2020roberta} as the backbone and fine-tune them on the evaluation datasets.
We compare with four deterministic embedding baselines: cross-entropy (CE) for classification tasks and mean squared error (MSE) for regression tasks, CE+CP~\cite{CECP}, CE/MSE+AT~\cite{CEAT}, and CE+SCL~\cite{CESCL}.
Besides, we compare \model{} with four probabilistic embedding models: VIB~\cite{dvib}, MINE-IB~\cite{belghazi2018mutual}, MEIB~\cite{an2023maximum}, and SPC~\cite{hu2024structured}.


\begin{table*}[t]
    \aboverulesep=0ex
    \belowrulesep=0ex
    \centering
    \setlength{\tabcolsep}{0.6mm}
    \caption{Classification evaluation (\%) results. The best results are bolded. w/o SE refers to \model{} without the proposed structural entropy based regularization loss.}\label{tab:classification results}
    \begin{tabular}{l|cccccccccc|c}
    \toprule
    Method & Emoji & Emotion & Hate & Irony & Offensive & Sentiment & Stance & ISEAR & MELD & GoEmotions  & Avg.\\
    \hline
    SVM & 29.30 & 64.70 & 36.70 & 61.70 & 52.30 & 62.90 & 67.30 & - & - & - & -\\
    FastText & 25.80 & 65.20 & 50.60 & 63.10 & 73.40 & 62.90 & 65.40 & - & - & - & -\\
    BiLSTM & 24.70 & 66.00 & 52.60 & 62.80 & 71.70 & 58.30 & 59.40 & - & - & - & -\\
    GPT-3.5 & 6.34\mysubscript{$\pm$0.01} & 73.23\mysubscript{$\pm$0.18} & 48.30\mysubscript{$\pm$0.11} & 66.81\mysubscript{$\pm$3.26} & 63.71\mysubscript{$\pm$0.13} & 40.40\mysubscript{$\pm$3.13} & 39.45\mysubscript{$\pm$0.10} & 67.22\mysubscript{$\pm$0.09} & 41.46\mysubscript{$\pm$0.11} & 25.21\mysubscript{$\pm$0.08} & 47.21\\
    \hline
    \multicolumn{12}{c}{\textit{BERT backbone}}\\
    \hline
    CE & 22.30\mysubscript{$\pm$0.60} & 76.05\mysubscript{$\pm$1.41} & 44.67\mysubscript{$\pm$1.78} & 59.38\mysubscript{$\pm$3.01} & 80.16\mysubscript{$\pm$1.26} & 70.54\mysubscript{$\pm$0.44} & 65.21\mysubscript{$\pm$0.71} & 67.17\mysubscript{$\pm$0.78} & 39.80\mysubscript{$\pm$0.84} & 46.29\mysubscript{$\pm$0.79} & 57.16\\
    CE+CP & 21.91\mysubscript{$\pm$0.71} & 76.28\mysubscript{$\pm$1.20} & 45.97\mysubscript{$\pm$2.93} & 64.06\mysubscript{$\pm$2.41} & 78.99\mysubscript{$\pm$1.57} & 70.68\mysubscript{$\pm$0.31} & 65.83\mysubscript{$\pm$0.39} & 67.20\mysubscript{$\pm$0.95} & 39.54\mysubscript{$\pm$1.61} & 46.39\mysubscript{$\pm$0.63} & 57.69\\
    CE+AT & 22.93\mysubscript{$\pm$0.70} & 75.08\mysubscript{$\pm$1.23} & 46.30\mysubscript{$\pm$3.61} & 64.23\mysubscript{$\pm$2.04} & 79.68\mysubscript{$\pm$1.59} & 70.55\mysubscript{$\pm$0.57} & 66.46\mysubscript{$\pm$1.13} &  65.70\mysubscript{$\pm$0.69} & 39.84\mysubscript{$\pm$0.38} & 47.37\mysubscript{$\pm$0.54} & 57.81\\
    CE+SCL & 21.72\mysubscript{$\pm$0.51} & 75.43\mysubscript{$\pm$1.37} & 45.86\mysubscript{$\pm$1.15} & 65.39\mysubscript{$\pm$2.46} & 80.20\mysubscript{$\pm$0.56} & 70.70\mysubscript{$\pm$0.79} & 65.34\mysubscript{$\pm$0.60} & 67.54\mysubscript{$\pm$0.64} & 40.00\mysubscript{$\pm$1.96} & 46.50\mysubscript{$\pm$0.46} & 57.87\\
    VIB & 21.31\mysubscript{$\pm$0.62} & 77.37\mysubscript{$\pm$0.71} & 45.99\mysubscript{$\pm$1.93} & 63.82\mysubscript{$\pm$1.00} & 80.37\mysubscript{$\pm$1.11} & 70.39\mysubscript{$\pm$0.31} & 65.43\mysubscript{$\pm$0.60} & 67.24\mysubscript{$\pm$0.57} & 38.52\mysubscript{$\pm$0.51} & 45.89\mysubscript{$\pm$1.10} & 57.63\\
    MINE-IB & 21.29\mysubscript{$\pm$0.31} & 76.60\mysubscript{$\pm$0.41} & 47.64\mysubscript{$\pm$2.11} & 65.86\mysubscript{$\pm$2.57} & 78.67\mysubscript{$\pm$2.28} & 69.85\mysubscript{$\pm$0.54} & 65.35\mysubscript{$\pm$0.88} & 67.62\mysubscript{$\pm$0.40} & 41.23\mysubscript{$\pm$0.67} & 46.87\mysubscript{$\pm$0.42} & 58.10\\
    MEIB & 21.87\mysubscript{$\pm$0.73} & 76.70\mysubscript{$\pm$0.82} & 48.27\mysubscript{$\pm$1.72} & 65.87\mysubscript{$\pm$2.14} & 80.49\mysubscript{$\pm$0.81} & 70.55\mysubscript{$\pm$0.57} & 65.59\mysubscript{$\pm$1.58} & 67.44\mysubscript{$\pm$0.50} & 39.30\mysubscript{$\pm$0.61} & 46.26\mysubscript{$\pm$0.81} & 58.23\\
    SPC & 24.19\mysubscript{$\pm$1.55} & 77.15\mysubscript{$\pm$0.73} & 57.48\mysubscript{$\pm$2.99} & 65.85\mysubscript{$\pm$1.07} & 80.65\mysubscript{$\pm$0.78} & 70.74\mysubscript{$\pm$0.12} & 67.17\mysubscript{$\pm$1.08} & 68.94\mysubscript{$\pm$0.35} & 42.68\mysubscript{$\pm$0.94} & 47.62\mysubscript{$\pm$1.38} & 60.25\\
    \textbf{\model{}} & 24.85\mysubscript{$\pm$0.31} & 78.58\mysubscript{$\pm$0.25} & 62.44\mysubscript{$\pm$2.08} & 69.56\mysubscript{$\pm$1.14} & 82.14\mysubscript{$\pm$0.59} & 71.35\mysubscript{$\pm$0.21} & 69.25\mysubscript{$\pm$0.78} & 69.77\mysubscript{$\pm$0.26} & 43.23\mysubscript{$\pm$0.71} & 51.16\mysubscript{$\pm$0.35} & 62.23\\
    - w/o SE & 22.49\mysubscript{$\pm$0.43} & 76.63\mysubscript{$\pm$1.07} & 56.54\mysubscript{$\pm$0.68} & 67.10\mysubscript{$\pm$0.54} & 80.31\mysubscript{$\pm$0.79} & 70.57\mysubscript{$\pm$0.54} & 66.84\mysubscript{$\pm$0.83} & 68.69\mysubscript{$\pm$0.14} & 42.31\mysubscript{$\pm$0.39} & 46.68\mysubscript{$\pm$0.32} & 59.82\\
    \hline
    \multicolumn{12}{c}{\textit{RoBERTa backbone}}\\
    \hline
    CE & 30.25\mysubscript{±1.32} & 77.41\mysubscript{±1.33} & 45.49\mysubscript{±4.70} & 57.99\mysubscript{±4.96} & 78.74\mysubscript{±2.20} & 71.80\mysubscript{±0.93} & 66.78\mysubscript{±1.34} & 70.00\mysubscript{±0.45} & 39.23\mysubscript{±0.41} & 46.64\mysubscript{±1.15} & 58.43\\
    CE+CP & 31.12\mysubscript{±0.84} & 77.54\mysubscript{±0.70} & 48.59\mysubscript{±3.28} & 58.75\mysubscript{±6.19} & 79.50\mysubscript{±0.98} & 72.82\mysubscript{±0.29} & 66.89\mysubscript{±1.67} & 70.58\mysubscript{±0.71} & 40.74\mysubscript{±0.89} & 47.98\mysubscript{±0.65} & 59.45\\
    CE+AT & 32.00\mysubscript{$\pm$0.93} & 77.30\mysubscript{$\pm$1.07} & 44.71\mysubscript{$\pm$4.76} & 60.17\mysubscript{$\pm$3.17} & 79.81\mysubscript{$\pm$1.11} & 72.51\mysubscript{$\pm$0.44} & 67.81\mysubscript{$\pm$0.95} & 70.97\mysubscript{$\pm$0.68} & 40.10\mysubscript{$\pm$0.60} & 47.89\mysubscript{$\pm$1.21} & 59.33\\
    CE+SCL & 31.09\mysubscript{$\pm$1.85} & 76.98\mysubscript{$\pm$2.02} & 49.51\mysubscript{$\pm$2.86} & 60.71\mysubscript{$\pm$4.23} & 80.39\mysubscript{$\pm$0.83} & 73.16\mysubscript{$\pm$0.44} & 66.73\mysubscript{$\pm$1.54} & 70.26\mysubscript{$\pm$0.45} & 40.64\mysubscript{$\pm$1.02} & 47.87\mysubscript{$\pm$0.86} & 59.72\\
    VIB & 29.71\mysubscript{$\pm$0.79} & 77.99\mysubscript{$\pm$0.86} & 49.39\mysubscript{$\pm$3.08} & 59.93\mysubscript{$\pm$4.57} & 79.63\mysubscript{$\pm$0.66} & 72.81\mysubscript{$\pm$0.39} & 68.40\mysubscript{$\pm$0.52} & 70.74\mysubscript{$\pm$0.44} & 38.94\mysubscript{$\pm$0.55} & 46.23\mysubscript{$\pm$0.18} & 59.38\\
    MINE-IB & 31.70\mysubscript{$\pm$0.45} & 78.79\mysubscript{$\pm$0.58} & 46.39\mysubscript{$\pm$2.82} & 57.39\mysubscript{$\pm$8.27} & 79.76\mysubscript{$\pm$0.67} & 72.85\mysubscript{$\pm$0.56} & 67.27\mysubscript{$\pm$1.00} & 70.15\mysubscript{$\pm$0.58} & 41.80\mysubscript{$\pm$2.14} & 48.88\mysubscript{$\pm$1.04} & 59.50\\
    MEIB & 29.94\mysubscript{$\pm$1.30} & 78.73\mysubscript{$\pm$0.90} & 49.34\mysubscript{$\pm$2.42} & 60.54\mysubscript{$\pm$2.70} & 79.68\mysubscript{$\pm$0.98} & 72.78\mysubscript{$\pm$0.29} & 67.89\mysubscript{$\pm$1.70} & 70.86\mysubscript{$\pm$0.61} & 39.00\mysubscript{$\pm$0.37} & 47.18\mysubscript{$\pm$1.15} & 59.59\\
    SPC & 32.54\mysubscript{$\pm$0.48} & 79.01\mysubscript{$\pm$0.61} & 59.80\mysubscript{$\pm$1.32} & 65.31\mysubscript{$\pm$1.91} & 80.98\mysubscript{$\pm$1.36} & 72.96\mysubscript{$\pm$0.22} & 69.02\mysubscript{$\pm$0.63} & 71.01\mysubscript{$\pm$0.59} & 43.99\mysubscript{$\pm$0.29} & 50.04\mysubscript{$\pm$0.60} & 62.47\\
    \textbf{\model{}} & \textbf{32.90\mysubscript{$\pm$0.22}} & \textbf{79.82\mysubscript{$\pm$0.54}} & \textbf{63.41\mysubscript{$\pm$1.27}} & \textbf{70.02\mysubscript{$\pm$1.22}} & \textbf{82.09\mysubscript{$\pm$0.46}} & 
    \textbf{73.18\mysubscript{$\pm$0.34}} & \textbf{70.33\mysubscript{$\pm$0.53}} & \textbf{71.92\mysubscript{$\pm$0.19}} & \textbf{44.64\mysubscript{$\pm$0.42}} & \textbf{51.55\mysubscript{$\pm$0.83}} & \textbf{63.99}\\
    - w/o SE & 31.05\mysubscript{$\pm$0.63} & 79.25\mysubscript{$\pm$0.33} & 57.13\mysubscript{$\pm$5.10} & 67.20\mysubscript{$\pm$0.86} & 80.74\mysubscript{$\pm$0.83} & 72.73\mysubscript{$\pm$0.19} & 69.06\mysubscript{$\pm$0.41} & 71.11\mysubscript{$\pm$0.92} & 43.23\mysubscript{$\pm$1.03} & 48.52\mysubscript{$\pm$0.86} & 62.00\\
    \bottomrule
    \end{tabular}
\end{table*}

\begin{table}[t]
    \aboverulesep=0ex
    \belowrulesep=0ex
    \centering
    \setlength{\tabcolsep}{0.5mm}
    \caption{Regression evaluation (\%) results with the RoBERTa backbone. The best results are bolded, and the second-best results are underlined. w/o soft refers to \model{} without the soft label and probabilistic encoding tree.}\label{tab:regression results}
    \begin{tabular}{l|cccc|c}
    \toprule
    \multirow{2}{*}{Method} & \multicolumn{2}{c}{STS-B} & \multicolumn{2}{c|}{Claire} & \multirow{2}{*}{Avg.}\\
    \cline{2-5} & Spearman & Pearson & Spearman & Pearson &\\
    \hline
    MSE & 88.33\mysubscript{$\pm$0.32} & 88.80\mysubscript{$\pm$0.36} & 50.37\mysubscript{$\pm$5.90} & 49.10\mysubscript{$\pm$5.74} & 69.15\\
    MSE+AT & 88.40\mysubscript{$\pm$0.50} & 89.01\mysubscript{$\pm$0.37} & 53.09\mysubscript{$\pm$0.64} & 51.87\mysubscript{$\pm$0.65} & 70.59\\
    VIB & 88.45\mysubscript{$\pm$0.50} & 89.01\mysubscript{$\pm$0.40} & 52.86\mysubscript{$\pm$0.88} & 51.66\mysubscript{$\pm$0.78} & 70.49\\
    MEIB & 88.61\mysubscript{$\pm$0.14} & 89.13\mysubscript{$\pm$0.17} & 52.85\mysubscript{$\pm$0.72} & 51.39\mysubscript{$\pm$0.81} & 70.50\\
    SPC & 88.71\mysubscript{$\pm$0.19} & 89.31\mysubscript{$\pm$0.24} & 53.11\mysubscript{$\pm$0.95} & 52.21\mysubscript{$\pm$0.81} & 70.84\\
    \textbf{\model{}} & \textbf{89.10\mysubscript{$\pm$0.29}} & \textbf{89.64\mysubscript{$\pm$0.20}} & \textbf{54.66\mysubscript{$\pm$0.69}} & \textbf{53.81\mysubscript{$\pm$0.84}} & \textbf{71.80}\\
    - w/o soft & 88.90\mysubscript{$\pm$0.29} & 89.27\mysubscript{$\pm$0.31} & 53.65\mysubscript{$\pm$0.64} & 52.85\mysubscript{$\pm$0.44} & 71.17\\
    \bottomrule
    \end{tabular}
\end{table}

\subsubsection{Parameter Settings} 
The training epoch number is 20, and the maximum patience for early stopping is 5 epochs.
The learning rate is 5e-5 in all datasets.
A linear learning rate warm-up is applied over the first 10\% of the training data.
The batch size is uniformly set to 128.
The trade-off parameter $\beta$ and the weight parameter $\gamma$ are searched from $\{1e-2, 1e-1, 1, 10\}$.
We set the class number $r=5$ for the STS-B dataset and $r=4$ for the Claire dataset, as their labels range from 0–5 and 1–5, respectively.
All experiments are conducted on two NVIDIA RTX A6000 GPUs.

\subsection{Evaluations}
\subsubsection{Classification Tasks}
We conduct comparative experiments with the baselines on 10 classification datasets and report the results in Table~\ref{tab:classification results}.
Both on the BERT backbone and the RoBERTa backbone, \model{} outperforms all other baselines with 2.02\%-5.07\% and 1.52\%-5.56\% average metric improvements, respectively.
Compared to SPC, which is also an encoder-only architecture-based probability coding model, \model{} still shows superior performance across all datasets.
These experimental results demonstrate the effectiveness of our proposed structural entropy based regularization loss $\mathcal{L}_{SE}$.
The most notable enhancement occurs in the Hate and the Irony dataset, where \model{} with the RoBERTa backbone surpasses all baselines with improvements ranging from 3.61\% to 26.71\% and 4.71\% to 12.63\%, respectively.
As the Hate datasets exhibit topic imbalance between the train and test sets, and the Irony dataset has higher requirements on language understanding because the semantics of ironic text are subtle compared to non-ironic text, the superior performance also demonstrates the better generalization capability of \model{}.

\begin{table*}[t]
    \aboverulesep=0ex
    \belowrulesep=0ex
    \centering
    \setlength{\tabcolsep}{0.3mm}
    \caption{Robustness analysis evaluation (\%) results against different noise rates. The best results are bolded.}\label{tab:noise results}
    \begin{tabular}{l|c|cccccccccc|c}
    \toprule
    Method & Noisy & Emoji & Emotion & Hate & Irony & Offensive & Sentiment & Stance & ISEAR & MELD & GoEmotions  & Avg.\\
    \hline
    CE & 10\% & 30.66\mysubscript{$\pm$0.89} & 78.15\mysubscript{$\pm$0.88} & 47.06\mysubscript{$\pm$5.40} & 56.90\mysubscript{$\pm$4.58} & 79.46\mysubscript{$\pm$0.80} & 72.36\mysubscript{$\pm$0.74} & 67.39\mysubscript{$\pm$1.86} & 70.40\mysubscript{$\pm$0.97} & 42.01\mysubscript{$\pm$1.94} & 47.85\mysubscript{$\pm$1.08} & 59.22\\
    VIB & 10\% & 30.74\mysubscript{$\pm$0.48} & 77.78\mysubscript{$\pm$2.05} & 47.64\mysubscript{$\pm$1.57} & 58.66\mysubscript{$\pm$10.60} & 79.96\mysubscript{$\pm$0.73} & 72.13\mysubscript{$\pm$0.54} & 67.54\mysubscript{$\pm$1.20} & 70.85\mysubscript{$\pm$0.33} & 38.63\mysubscript{$\pm$0.89} & 47.30\mysubscript{$\pm$1.65} & 59.12\\
    MINE-IB & 10\% & 31.14\mysubscript{$\pm$0.65} & 78.04\mysubscript{$\pm$1.03} & 47.19\mysubscript{$\pm$3.29} & 56.80\mysubscript{$\pm$8.63} & 78.36\mysubscript{$\pm$1.46} & 72.42\mysubscript{$\pm$0.47} & 67.16\mysubscript{$\pm$1.51} & 70.34\mysubscript{$\pm$0.44} & 42.32\mysubscript{$\pm$1.65} & 48.56\mysubscript{$\pm$1.41} & 59.23\\
    MEIB & 10\% & 31.02\mysubscript{$\pm$0.47} & 78.94\mysubscript{$\pm$0.46} & 49.28\mysubscript{$\pm$4.58} & 57.21\mysubscript{$\pm$8.07} & 80.19\mysubscript{$\pm$0.83} & 72.09\mysubscript{$\pm$0.68} & 68.26\mysubscript{$\pm$0.68} & 70.85\mysubscript{$\pm$0.38} & 38.67\mysubscript{$\pm$0.97} & 46.93\mysubscript{$\pm$1.06} & 59.34\\
    SPC & 10\% & 32.25\mysubscript{$\pm$0.69} & 78.88\mysubscript{$\pm$0.47} & 56.13\mysubscript{$\pm$5.36} & 58.88\mysubscript{$\pm$4.94} & 80.14\mysubscript{$\pm$0.28} & 72.76\mysubscript{$\pm$0.06} & 68.57\mysubscript{$\pm$1.01} & 71.10\mysubscript{$\pm$0.62} & 43.90\mysubscript{$\pm$1.13} & 49.32\mysubscript{$\pm$1.22} & 61.19\\
    \textbf{\model{}} & 10\% & \textbf{32.92\mysubscript{$\pm$0.39}} & \textbf{79.17\mysubscript{$\pm$0.56}} & \textbf{60.93\mysubscript{$\pm$1.96}} & \textbf{69.86\mysubscript{$\pm$1.33}} & \textbf{81.33\mysubscript{$\pm$0.21}} & \textbf{72.99\mysubscript{$\pm$0.18}} & \textbf{69.33\mysubscript{$\pm$0.99}} & \textbf{71.61\mysubscript{$\pm$0.35}} & \textbf{44.57\mysubscript{$\pm$0.19}} & \textbf{51.53\mysubscript{$\pm$0.64}} & \textbf{63.42} \\
    \hline
    CE & 20\% & 31.96\mysubscript{$\pm$0.88} & 77.01\mysubscript{$\pm$1.51} & 49.12\mysubscript{$\pm$0.72} & 60.82\mysubscript{$\pm$3.56} & 79.54\mysubscript{$\pm$1.64} & 72.06\mysubscript{$\pm$0.63} & 68.49\mysubscript{$\pm$1.20} & 70.32\mysubscript{$\pm$0.26} & 40.16\mysubscript{$\pm$1.94} & 47.78\mysubscript{$\pm$0.84} & 59.73\\
    VIB & 20\% & 30.46\mysubscript{$\pm$0.59} & 79.00\mysubscript{$\pm$0.49} & 47.91\mysubscript{$\pm$2.20} & 60.67\mysubscript{$\pm$4.82} & 79.15\mysubscript{$\pm$1.22} & 72.26\mysubscript{$\pm$0.29} & 66.83\mysubscript{$\pm$0.52} & 71.02\mysubscript{$\pm$0.25} & 39.33\mysubscript{$\pm$1.47} & 47.83\mysubscript{$\pm$1.38} & 59.45\\
    MINE-IB & 20\% & 30.31\mysubscript{$\pm$0.97} & 77.84\mysubscript{$\pm$0.98} & 46.23\mysubscript{$\pm$3.23} & 57.43\mysubscript{$\pm$8.41} & 78.65\mysubscript{$\pm$0.91} & 72.02\mysubscript{$\pm$0.83} & 66.83\mysubscript{$\pm$1.82} & 69.26\mysubscript{$\pm$0.52} & 42.31\mysubscript{$\pm$1.58} & 47.55\mysubscript{$\pm$0.99} & 58.84\\
    MEIB & 20\% & 30.84\mysubscript{$\pm$0.75} & 78.38\mysubscript{$\pm$0.88} & 50.02\mysubscript{$\pm$5.18} & 55.12\mysubscript{$\pm$7.07} & 78.17\mysubscript{$\pm$2.55} & 71.63\mysubscript{$\pm$1.11} & 68.05\mysubscript{$\pm$0.81} & 70.68\mysubscript{$\pm$0.38} & 39.09\mysubscript{$\pm$0.87} & 47.29\mysubscript{$\pm$1.22} & 58.93\\
    SPC & 20\% & 32.51\mysubscript{$\pm$0.83} & 77.97\mysubscript{$\pm$1.12} & 55.41\mysubscript{$\pm$6.00} & 66.40\mysubscript{$\pm$4.26} & 80.33\mysubscript{$\pm$0.48} & 72.50\mysubscript{$\pm$0.55} & 68.89\mysubscript{$\pm$1.60} & 71.10\mysubscript{$\pm$0.39} & 43.96\mysubscript{$\pm$0.50} & 49.04\mysubscript{$\pm$0.42} & 61.81\\
    \textbf{\model{}} & 20\% & \textbf{33.04\mysubscript{$\pm$0.19}} & \textbf{79.55\mysubscript{$\pm$0.42}} & \textbf{60.30\mysubscript{$\pm$1.80}} & \textbf{69.61\mysubscript{$\pm$1.51}} & \textbf{81.66\mysubscript{$\pm$0.44}} & \textbf{72.97\mysubscript{$\pm$0.30}} & \textbf{69.81\mysubscript{$\pm$0.64}} & \textbf{71.68\mysubscript{$\pm$0.26}} & \textbf{44.50\mysubscript{$\pm$0.86}} & \textbf{51.64\mysubscript{$\pm$0.42}} & \textbf{63.48}\\
    \hline
    CE & 30\% & 31.82\mysubscript{$\pm$0.75} & 77.61\mysubscript{$\pm$0.90} & 50.69\mysubscript{$\pm$2.80} & 58.90\mysubscript{$\pm$11.45} & 78.11\mysubscript{$\pm$2.07} & 70.15\mysubscript{$\pm$0.50} & 69.07\mysubscript{$\pm$1.07} & 70.74\mysubscript{$\pm$0.56} & 40.61\mysubscript{$\pm$2.06} & 47.76\mysubscript{$\pm$2.29} & 59.55\\
    VIB & 30\% & 30.85\mysubscript{$\pm$0.53} & 78.23\mysubscript{$\pm$0.79} & 48.22\mysubscript{$\pm$1.97} & 58.81\mysubscript{$\pm$8.84} & 79.38\mysubscript{$\pm$0.62} & 72.15\mysubscript{$\pm$0.52} & 67.59\mysubscript{$\pm$0.93} & 70.27\mysubscript{$\pm$0.74} & 38.71\mysubscript{$\pm$1.19} & 47.16\mysubscript{$\pm$1.32} & 59.14\\
    MINE-IB & 30\% & 30.12\mysubscript{$\pm$0.79} & 77.82\mysubscript{$\pm$1.24} & 46.05\mysubscript{$\pm$3.94} & 56.02\mysubscript{$\pm$7.24} & 78.26\mysubscript{$\pm$1.58} & 72.23\mysubscript{$\pm$0.74} & 65.56\mysubscript{$\pm$2.67} & 69.55\mysubscript{$\pm$0.92} & 39.46\mysubscript{$\pm$1.82} & 46.71\mysubscript{$\pm$1.87} & 58.18\\
    MEIB & 30\% & 30.74\mysubscript{$\pm$0.87} & 77.99\mysubscript{$\pm$0.69} & 49.98\mysubscript{$\pm$4.00} & 57.57\mysubscript{$\pm$5.19} & 72.53\mysubscript{$\pm$5.53} & 71.83\mysubscript{$\pm$0.40} & 67.88\mysubscript{$\pm$0.68} & 69.86\mysubscript{$\pm$1.24} & 39.39\mysubscript{$\pm$1.06} & 47.43\mysubscript{$\pm$1.52} & 58.52\\
    SPC & 30\% & 32.27\mysubscript{$\pm$0.48} & 78.13\mysubscript{$\pm$1.13} & 56.04\mysubscript{$\pm$7.44} & 59.27\mysubscript{$\pm$8.56} & 80.32\mysubscript{$\pm$0.53} & 72.44\mysubscript{$\pm$0.36} & 69.77\mysubscript{$\pm$0.93} & 70.91\mysubscript{$\pm$0.30} & 43.29\mysubscript{$\pm$0.53} & 49.82\mysubscript{$\pm$2.55} & 61.23\\
    \textbf{\model{}} & 30\% & \textbf{32.80\mysubscript{$\pm$0.09}} & \textbf{79.49\mysubscript{$\pm$0.63}} & \textbf{60.19\mysubscript{$\pm$1.91}} & \textbf{68.74\mysubscript{$\pm$1.83}} & \textbf{81.55\mysubscript{$\pm$0.44}} & \textbf{72.73\mysubscript{$\pm$0.19}} & \textbf{69.79\mysubscript{$\pm$0.54}} & \textbf{71.57\mysubscript{$\pm$0.49}} & \textbf{44.89\mysubscript{$\pm$0.71}} & \textbf{51.49\mysubscript{$\pm$0.53}} & \textbf{63.32}\\
    
    \bottomrule
    \end{tabular}
\end{table*}

\begin{figure*}[h!]
    \centering
    \includegraphics[width=0.90\linewidth]{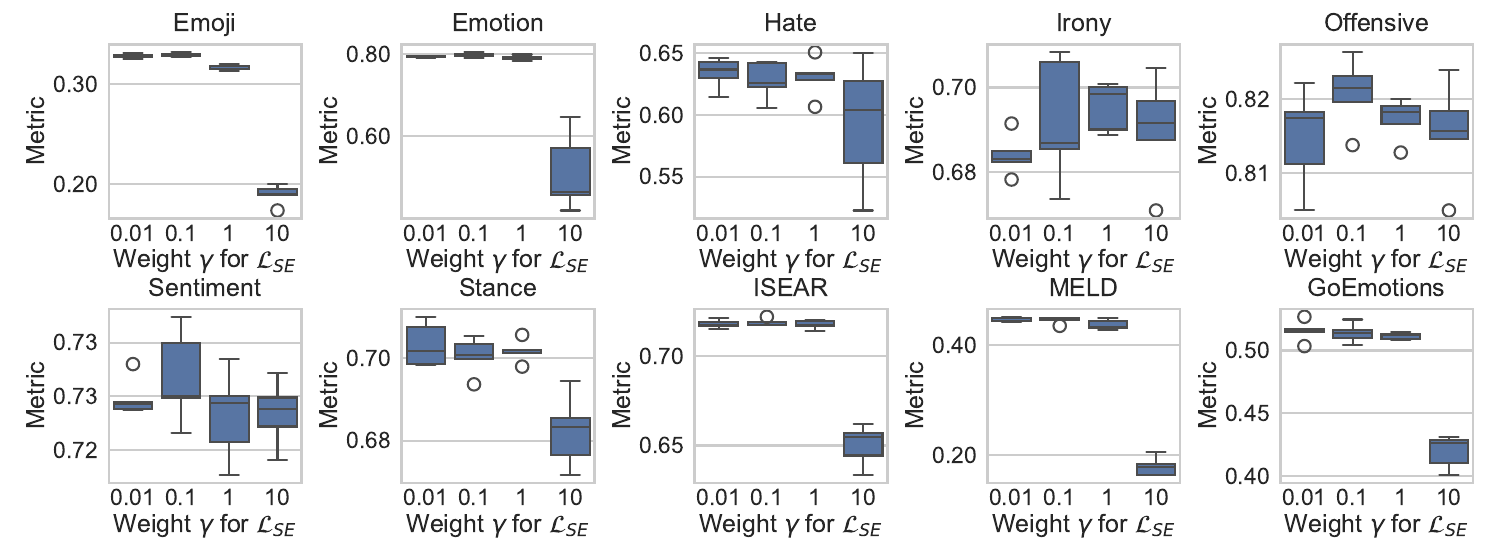}
    \caption{The impact of the weight parameter $\gamma$ on the regularization loss $\mathcal{L}_{SE}$.}
    \label{fig:hyperparameter}
\end{figure*}
We also conduct an ablation study on \model{} w/o SE model, disable the proposed $\mathcal{L}_{SE}$, and report the results in Table~\ref{tab:classification results}.
It is noteworthy that, despite \model{} w/o SE model being the same as SPC w/o S model~\cite{hu2024structured}, we report our experimental results as the hyperparameters and experiment environments differ.
The absence of $\mathcal{L}_{SE}$ leads to a performance decrease of an average of 2.41\% and 1.99\% on the BERT and RoBERTa backbones, respectively.
This indicates the effectiveness of our proposed structural entropy based regularization loss.

\subsubsection{Regression Tasks}
We experiment with regression tasks on STS-B and Claire datasets and report Spearman and Pearson correlation coefficients results in Table~\ref{tab:regression results}.
All methods use RoBERTa as the backbone.
\model{} outperforms all other baselines across all datasets, with an average of 0.96\%-2.65\% performance improvement.
This demonstrates the effectiveness of \model{} on the regression tasks.
To better understand our proposed probabilistic encoding tree, we conduct an ablation study by removing the soft label and probabilistic encoding tree.
Instead, we assign each sample to the class with the closest distance to the class bin center and use the normal encoding tree to calculate $\mathcal{L}_{SE}$.
As shown in Table~\ref{tab:regression results}, \model{} outperforms \model{} without the soft label and probabilistic encoding tree.
This proves the information loss of directly discretizing regression labels and also indicates the effectiveness of our proposed method of softening and probabilistic encoding in the regression tasks.

\begin{figure*}[t]
    \centering
    \includegraphics[width=0.89\linewidth]{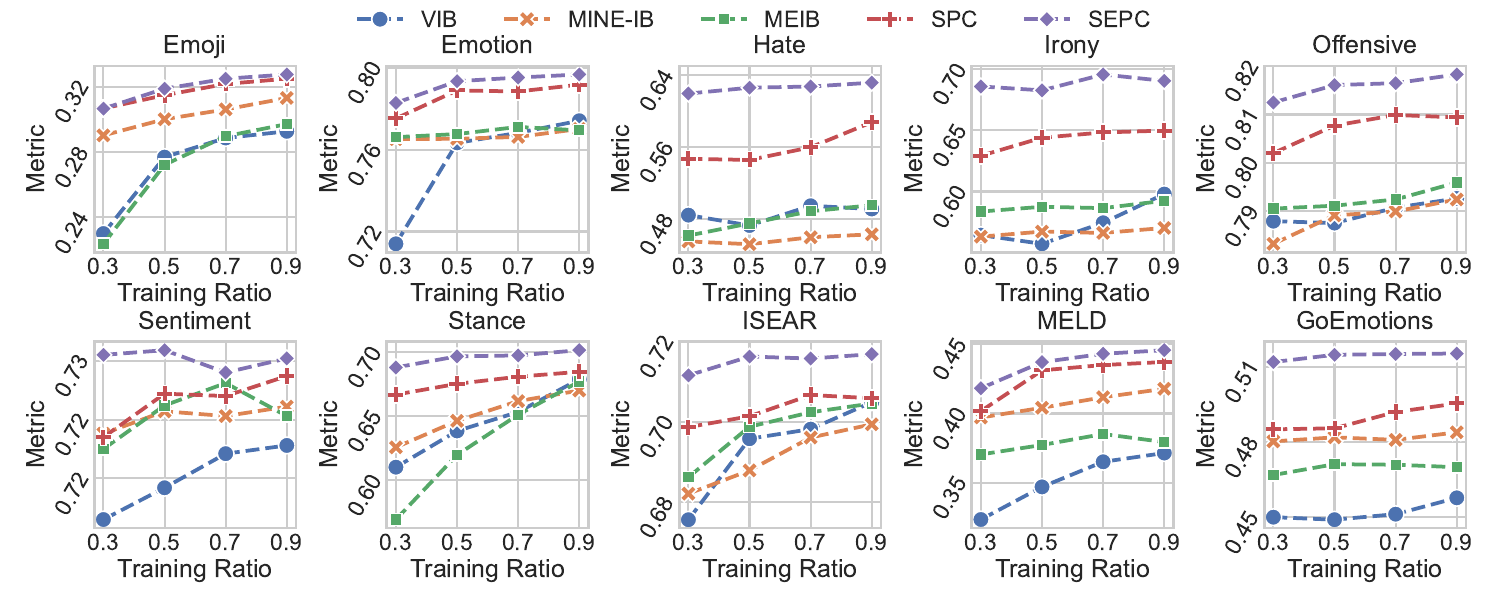}
    \caption{Results of different models with different ratios of the training set.}
    \label{fig:generalization}
\end{figure*}

\subsubsection{Robustness Analysis}
To evaluate the robustness of \model{}, we introduce noise by randomly flipping 10\%, 20\%, and 30\% of the labels in the training datasets to any class with the same probability.
The experimental results are reported in Table~\ref{tab:noise results}.
\model{} shows superior performance across all noise rate settings and all datasets compared to baselines.
Specifically, it outperforms all baselines with average improvements of 2.23\%-4.30\%, 1.67\%-4.64\%, and 2.09\%-5.14\% under 10\%, 20\%, and 30\% noise rates, respectively.
Besides, when the noise rate increases from 20\% to 30\%, \model{} exhibits a minimal average performance decrease.
This experiment demonstrates that \model{} has better robustness when handling noise and data unreliability.

\subsubsection{Hyperparameter Sensitivity Analysis}
We evaluate the impact of the newly introduced weight hyperparameter $\gamma$ for regularization loss $\mathcal{L}_{SE}$ on ten classification datasets and illustrate the results in Figure~\ref{fig:hyperparameter}.
A lower regularization weight is preferred in most datasets.
A too-large weight, $\gamma = 10$, generally leads to a noticeable performance decrement and higher variance.

\subsubsection{Generalization Analysis}
We conduct experiments under limited training data conditions to better evaluate the generalization capability of \model{}.
In detail, we randomly select 90\%, 70\%, 50\%, and 30\% of the training data during the model training period and compare the performance of \model{} on the test set with other probabilistic coding models.
The experimental results are illustrated in Figure~\ref{fig:generalization}.
\model{} outperforms all other baselines across all datasets under different percentages of the training set.
The superior performance under the limited training data demonstrates the generalization capability of \model{}.

\subsubsection{Visualization}
\begin{figure}[t]
    \centering
    \includegraphics[width=0.90\linewidth]{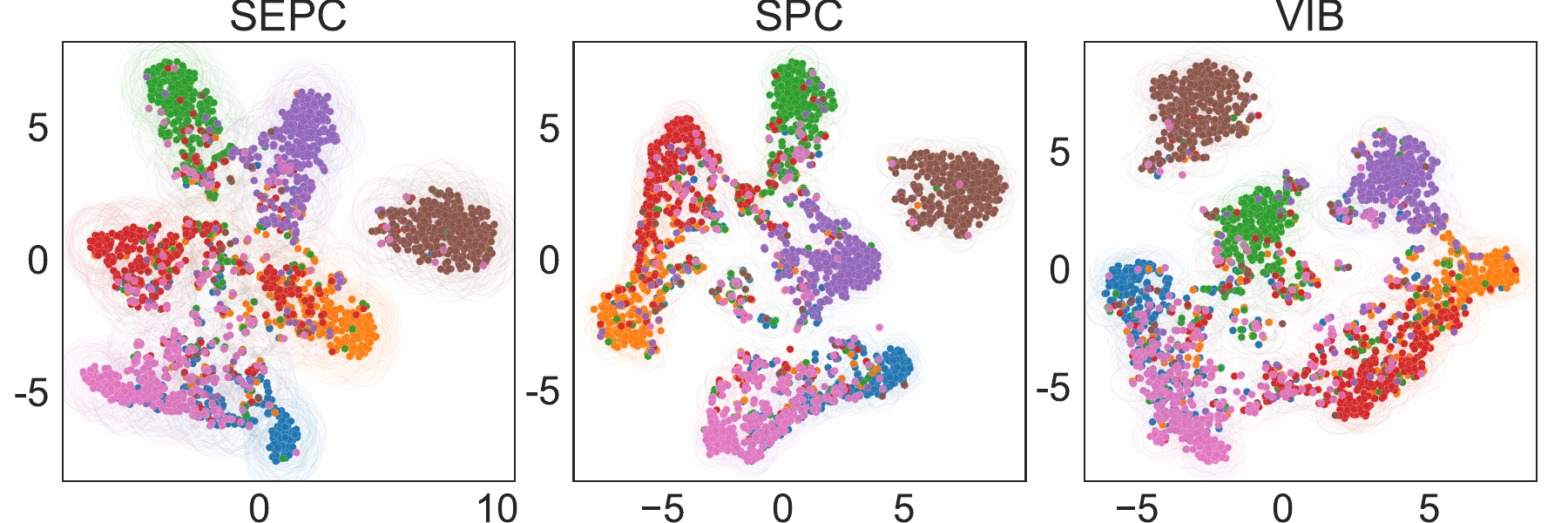}
    \caption{Visualization of embeddings. The circle represents the standard deviation of the probabilistic embeddings.}
    \label{fig:tsne}
\end{figure}
We visualize the embeddings of \model{}, VIB, and SPC on the ISEAR dataset to intuitively showcase the advantages of \model{}'s learned embeddings.
As shown in Figure~\ref{fig:tsne}, the embeddings of \model{} are more discriminative.
Additionally, the embedding distribution of \model{} has a larger standard deviation, thus occupying a larger embedding space.
This results in better generalization capability.
\section{Related Work}\label{sec:related work}
\subsubsection{Probabilistic Embedding}
Compared to deterministic embedding~\cite{Dong2024, xu2024sctnet}, probabilistic embedding learns a probabilistic distribution for each input, effectively capturing data uncertainty and complexity, and thus better handling noise and outliers. 
The mainstream probabilistic embedding methods follow the Information Bottleneck (IB) principle~\cite{tishby2000information, tishby2015deep}, which seeks to discover compressed representations that retain the maximum amount of relevant information for the prediction task while eliminating as much irrelevant information as possible.
VIB~\cite{dvib} constrains the latent variable to follow the Gaussian distribution and utilizes Kullback-Leibler (KL) divergence between the learned distribution and the prior Gaussian distribution as the regularization loss.
Sparse IB~\cite{chalk2016relevant} replaces the prior Gaussian distribution of VIB with the Student-t distribution.
MINE-IB~\cite{belghazi2018mutual} is a mutual information neural estimation method with the IB principle, allowing for the tractable IB application in a continuous setting.
\citet{fischer2020conditional} proposes the conditional entropy bottleneck top improved robustness to adversarial examples.
MEIB~\cite{an2023maximum} utilizes maximum conditional entropy to serve as the bottleneck of IB.
SPC~\cite{hu2024structured} introduces an encoder-only framework, incorporating a class-level structured regularization loss.

\subsubsection{Structural Entropy}
Unlike early information entropy, such as Shannon entropy, which is defined by unstructured probability distributions, structural entropy~\cite{li2016structural} takes the hierarchical structural information of the input data into account.
It is gaining substantial traction and is widely used in graph structural learning~\cite{zou2023segsl}, node classification~\cite{duan2024structural}, social bot detection~\cite{pengunsupervised2024, zeng2024adversarial}, and deep clustering~\cite{sun2024lsenet}.
USER~\cite{wang2023user} proposes a structural entropy-based loss.
However, current works focus solely on minimizing structural entropy to maximize task-related information and are limited to classification tasks.

\section{Conclusion}\label{sec:conclusion}
In this paper, we propose \model{}, a structural entropy guided probabilistic coding model.
\model{} utilizes maximizing the structural entropy as the regularization loss, introducing the structural information into the optimization, and aims to separate the latent variables in the class space.
Additionally, we propose a probabilistic encoding tree and an effective method to utilize the structural entropy for regression tasks based on it.
Experiments on 12 datasets demonstrate the effectiveness, generalization capability, and robustness of \model{} in both classification and regression tasks.

\section*{Acknowledgments}
This work is supported by the National Key R\&D Program of China through grant 2022YFB3104703, NSFC through grants 62322202, 62432006, and 62476163, Local Science and Technology Development Fund of Hebei Province Guided by the Central Government of China through grant 246Z0102G, Guangdong Basic and Applied Basic Research Foundation through grant 2023B1515120020, Open Research Fund from Guangdong Laboratory of Artificial Intelligence and Digital Economy (SZ) under Grant No. GML-KF-24-08, and CCF-DiDi GAIA Collaborative Research Funds for Young Scholars.

\bibliography{aaai25}

\clearpage
\appendix

\end{document}